\documentclass{article}

    \usepackage[preprint]{neurips_2024}

\usepackage[utf8]{inputenc}
\usepackage[T1]{fontenc}
\usepackage{hyperref}
\usepackage{url}
\usepackage{booktabs}
\usepackage{amsfonts}
\usepackage{nicefrac}
\usepackage{microtype}
\usepackage{xcolor}

\usepackage{hyperref}
\hypersetup{
    colorlinks=true,
    linkcolor=blue,
    filecolor=blue,
    urlcolor=blue,
    citecolor=blue,
    }
\usepackage{graphicx}
\usepackage{array}
\usepackage{makecell}
\usepackage{longtable}
\usepackage{listings}
\usepackage{adjustbox}

\title{Prompt Repetition Improves Non-Reasoning LLMs}

\author{
  Yaniv Leviathan\thanks{Equal contribution.} \\
  Google Research\\
  \texttt{leviathan@google.com} \\
  \And
  Matan Kalman\footnotemark[1] \\
  Google Research \\
  \texttt{matank@google.com} \\
  \And
  Yossi Matias \\
  Google Research \\
  \texttt{yossi@google.com} \\
}

\begin{document}

\maketitle

\begin{abstract}

\end{abstract}
When not using reasoning, repeating the input prompt improves performance for popular models (Gemini, GPT, Claude, and Deepseek) without increasing the number of generated tokens or latency.

\section{Prompt Repetition}
\label{section:intro}

\begin{figure}[h]
  \centering
  \begin{adjustbox}{center}
    \includegraphics[width=\textwidth]{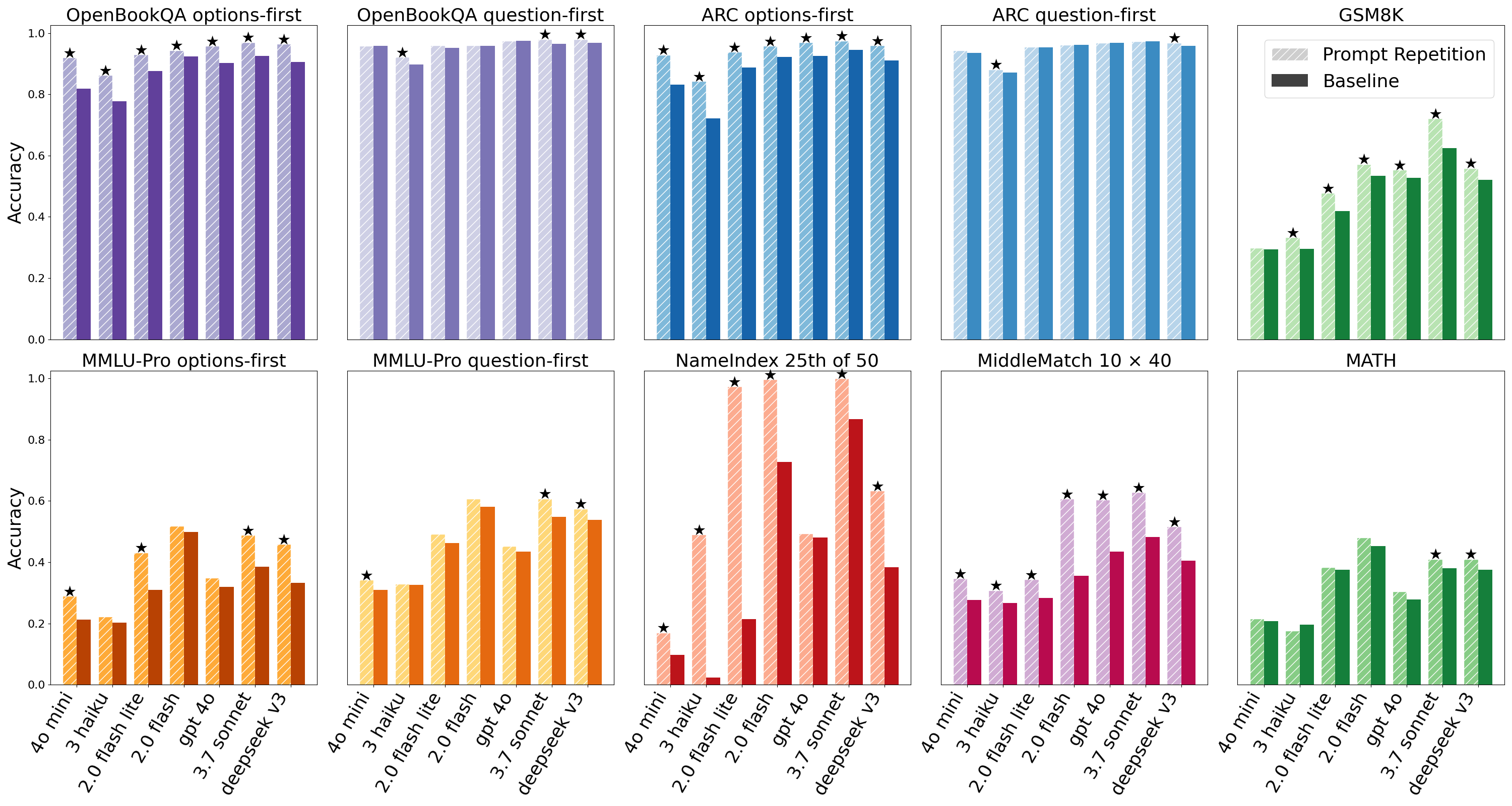}
  \end{adjustbox}
  \caption{Prompt repetition vs. baseline accuracy for popular LLMs and various benchmarks when asking the models not to reason. A star indicates a statistically significant win ($p_{value} < 0.1$ according to the McNemar test \cite{mcnemar1947note}). \textbf{Prompt repetition wins 47 out of 70 tests, with 0 losses}.}
  \label{fig:acc}
\end{figure}

LLMs are often trained as \emph{causal} language models, i.e. past tokens cannot attend to future tokens.
Therefore, the order of the tokens in a user's query can affect prediction performance.
For example, a query of the form ``\texttt{<CONTEXT> <QUESTION>}'' often performs differently from a query of
the form ``\texttt{<QUESTION> <CONTEXT>}'' (see \emph{options-first} vs. \emph{question-first} in Figure \ref{fig:acc}).
We propose to \emph{repeat the prompt}, i.e. transform the input from ``\colorbox{blue!10}{\texttt{<QUERY>}}'' to ``{\colorbox{blue!10}{\texttt{<QUERY><QUERY>}}}''. This enables each prompt token to attend to every other prompt token, addressing the above. When not using reasoning, \textbf{prompt repetition improves the performance of LLMs} (\mbox{Figure \ref{fig:acc}}) \textbf{without increasing the lengths of the generated outputs or latency} (Figures \ref{fig:ablations} and \ref{fig:ablations2}).

As further motivation, we observe that reasoning models trained with RL often learn to repeat (parts of) the user's request.
Prompt repetition is \emph{efficient}, moving the repetition to the parallelizable prefill stage. The number of generated tokens does not increase \footnote[1]{Unlike other prompting techniques, such as ``Think step by step'' \citep{lets_think_step_by_step_kojima2023largelanguagemodelszeroshot}, see Figures \ref{fig:ablations} and \ref{fig:ablations2}.}. Moreover, prompt repetition does not change the format of the generated outputs, keeping them interchangeable with those of the original prompts, enabling simple drop-in deployment in existing systems, and direct use by end-users.
When reasoning is enabled, prompt repetition is neutral to slightly positive (Figure \ref{fig:sbs}).

\section{Experiments}
\label{section:experiments}

We test \emph{prompt repetition} on a range of 7 popular models from leading LLM providers of varying sizes: Gemini 2.0 Flash and Gemini 2.0 Flash Lite \citep{team2023gemini}, GPT-4o-mini and GPT-4o \citep{openai2024gpt4ocard}, Claude 3 Haiku and Claude 3.7 Sonnet \citep{claude}, and Deepseek V3 \citep{deepseekai2025deepseekv3technicalreport}.
We ran all tests using each provider's official API in Feb and Mar 2025.

We test each of the models on a set of 7 benchmarks in several configurations: ARC (Challenge) \citep{arc_clark2018thinksolvedquestionanswering}, OpenBookQA \citep{openbookqa_mihaylov2018suitarmorconductelectricity}, GSM8K \citep{gsm8k_cobbe2021trainingverifierssolvemath}, MMLU-Pro \citep{wang2024mmluprorobustchallengingmultitask}, MATH \citep{hendrycksmath2021}, and 2 custom benchmarks: NameIndex and MiddleMatch (Appendix \ref{appendix-custom-tasks}).
For the multiple choice benchmarks (ARC, OpenBookQA, and MMLU-Pro) we report results both when placing the question first and the answer options later, as well as the other way around, the latter making the model process the options without seeing the question in context (unless using prompt repetition).

\paragraph{Accuracy.} Without reasoning, prompt repetition improves the accuracy of all tested LLMs and benchmarks (Figure \ref{fig:acc}).
We consider cases where one method is significantly better than the other according to the McNemar test \citep{mcnemar1947note} with $p_{value} < 0.1$ as wins.
With this criteria, \textbf{prompt repetition wins 47 out of 70 benchmark-model combinations, with 0 losses}.
Notably, performance is improved for all tested models.
As expected, we observe smaller improvements for the multiple-choice benchmarks with question-first, and larger improvements with options-first.
On the custom tasks of NameIndex and MiddleMatch we observe strong gains with prompt repetition for all models (for example, prompt repetition improves the accuracy of Gemini 2.0 Flash-Lite on NameIndex from 21.33\% to 97.33\%).
We also test a smaller set of benchmarks when encouraging thinking step-by-step (Figure \ref{fig:sbs}), where the results are neutral to slightly positive (5 wins, 1 loss, and 22 neutral), as expected (see Appendix \ref{appendix:with reasoning}).

\paragraph{Ablations and variations.} We compare prompt repetition to 2 additional variants: \emph{Prompt Repetition (Verbose)} and \emph{Prompt Repetition $\times 3$} (Appendix \ref{appendix-query-examples}). We observe that they perform similarly for most tasks and models (Figures \ref{fig:ablations} and \ref{fig:ablations2}), and sometimes outperform vanilla prompt repetition.
Notably, \emph{Prompt Repetition $\times 3$} often substantially outperforms vanilla prompt repetition (which itself substantially outperforms the baseline) on NameIndex and MiddleMatch.
It therefore seems worthwhile to further research variants.
To demonstrate that the gains are indeed due to repeating the prompt and not to simply increasing the length of the input, we also evaluate the \emph{Padding} method (Appendix \ref{appendix-query-examples}), which pads the inputs with periods (``.'') to the same length as prompt repetition, and, as expected, does not improve performance.

\paragraph{Efficiency.} For each model, prompting method, and dataset, we measure the average and median of the lengths of the generated outputs, as well as the empirical latency\footnote{We measure end-to-end latencies via the official API for each of the providers. These might be affected by e.g., network delays or transient loads. For fairness, we ran all requests to the same provider in a round-robin fashion. As can be seen in Figures \ref{fig:ablations} and \ref{fig:ablations2} the measured latencies are consistent with expectation, but given the above, should be taken with a grain of salt. Notably, the measured latencies for Deepseek are very high.}.
As expected, we observe similar latencies for all datasets and all tested models when reasoning is disabled. With reasoning enabled, all latencies (and the lengths of the generated outputs) are dramatically higher.
Either way, in all cases \textbf{prompt repetition and its variants do not increase the lengths of the generated outputs or the measured latencies} (Figures \ref{fig:ablations} and \ref{fig:ablations2}), the only exception being the Anthropic models (Claude Haiku and Sonnet) for very long requests (from the NameIndex or MiddleMatch datasets or from the repeat $\times 3$ variant) where the latencies increase (likely due to the prefill stage taking longer).

\section{Related Work}
\label{section:related work}

Many prompting techniques for LLMs have been suggested, notably Chain of Thought (CoT) prompting \citep{wei2023chainofthoughtpromptingelicitsreasoning} (which requires specific examples per task) and ``Think step by step'' \citep{lets_think_step_by_step_kojima2023largelanguagemodelszeroshot}, which achieves substantial improvements, but increases the lengths of the generated outputs and thus the latency and compute requirements (we show that it can be used in tandem with prompt repetition, yielding mostly neutral results). More recently and independently, \cite{shaier2024askingagainexploringllm} experimented with repeating just the question part of the prompt and found that it yields no gains, \cite{springer2024repetitionimproveslanguagemodel} showed that repeating the input twice yields better text embeddings, and \cite{xu2024rereadingimprovesreasoninglarge} showed that asking the model to re-read the question improves reasoning.

\section{Conclusion}
\label{section:conclusion}

We show that repeating the prompts consistently improves model performance for a range of models and benchmarks, when not using reasoning.
In addition, latency is not impacted\footnote{Prompt repetition can affect latency for long prompts, and might be impossible for very long ones.}, as only the parallelizable pre-fill stage is affected\footnote{This parallelization has a similar motivation to e.g., speculative decoding \citep{leviathan2022fastinferencetransformersspeculative}.}.
Prompt repetition does not change the lengths or formats of the generated outputs, and it might be a good default for many models and tasks, when reasoning is not used.

\paragraph{Future directions.} 

(1) Fine tune the model with repeated prompts;
(2) Train a reasoning model with prompt repetition to increase efficiency (the model might learn to avoid repeating the prompt);
(3) Periodically repeat the last generated tokens during generation, as well as explore applicability to multi-turn scenarios;
(4) Only keep the second repetition in the KV-cache (thus being completely performance neutral for the generation stage);
(5) Repeat only parts of the prompt (especially for longer prompts);
(6) Reorder the prompt, e.g. with a smaller model, instead of repeating everything;
(7) Applicability to non-text modalities (e.g. images);
(8) Further analyze different variants, e.g. when more than 2 repetitions might be advantageous;
(9) Further analyze the attention patterns due to repetition (in the same vein as done in \cite{xu2024rereadingimprovesreasoninglarge});
(10) Use repetitions in tandem with techniques like selective attention \citep{leviathan2024selectiveattentionimprovestransformer};
(11) Explore interactions with techniques such as Prefix LM \citep{raffel2023exploringlimitstransferlearning};
(12) Investigate when repetition is helpful and how token representations vary between repetitions;
(13) Explore promising variants (see Appendix~\ref{appendix:ablations and variations}).

\section*{Acknowledgements}

We’d like to extend a huge thank you to Raya Leviathan, Danny Lumen, Dani Valevski, Sergey Levi, the Theta Labs and Google Research teams, and our families for insightful feedback, ideas, suggestions, and support.

\bibliographystyle{plainnat}
\bibliography{bib}

\begin{thebibliography}{18}
\providecommand{\natexlab}[1]{#1}
\providecommand{\url}[1]{\texttt{#1}}
\expandafter\ifx\csname urlstyle\endcsname\relax
  \providecommand{\doi}[1]{doi: #1}\else
  \providecommand{\doi}{doi: \begingroup \urlstyle{rm}\Url}\fi

\bibitem[Anthropic(2024)]{claude}
Anthropic.
\newblock The claude 3 model family: Opus, sonnet, haiku.
\newblock 2024.
\newblock URL
  \url{https://www-cdn.anthropic.com/de8ba9b01c9ab7cbabf5c33b80b7bbc618857627/Model_Card_Claude_3.pdf}.

\bibitem[Clark et~al.(2018)Clark, Cowhey, Etzioni, Khot, Sabharwal, Schoenick,
  and Tafjord]{arc_clark2018thinksolvedquestionanswering}
Peter Clark, Isaac Cowhey, Oren Etzioni, Tushar Khot, Ashish Sabharwal, Carissa
  Schoenick, and Oyvind Tafjord.
\newblock Think you have solved question answering? try arc, the ai2 reasoning
  challenge, 2018.
\newblock URL \url{https://arxiv.org/abs/1803.05457}.

\bibitem[Cobbe et~al.(2021)Cobbe, Kosaraju, Bavarian, Chen, Jun, Kaiser,
  Plappert, Tworek, Hilton, Nakano, Hesse, and
  Schulman]{gsm8k_cobbe2021trainingverifierssolvemath}
Karl Cobbe, Vineet Kosaraju, Mohammad Bavarian, Mark Chen, Heewoo Jun, Lukasz
  Kaiser, Matthias Plappert, Jerry Tworek, Jacob Hilton, Reiichiro Nakano,
  Christopher Hesse, and John Schulman.
\newblock Training verifiers to solve math word problems, 2021.
\newblock URL \url{https://arxiv.org/abs/2110.14168}.

\bibitem[DeepSeek-AI(2025)]{deepseekai2025deepseekv3technicalreport}
DeepSeek-AI.
\newblock Deepseek-v3 technical report, 2025.
\newblock URL \url{https://arxiv.org/abs/2412.19437}.

\bibitem[{Gemini Team Google}(2023)]{team2023gemini}
{Gemini Team Google}.
\newblock Gemini: A family of highly capable multimodal models.
\newblock \emph{arXiv preprint arXiv:2312.11805}, 2023.

\bibitem[Hendrycks et~al.(2021)Hendrycks, Burns, Kadavath, Arora, Basart, Tang,
  Song, and Steinhardt]{hendrycksmath2021}
Dan Hendrycks, Collin Burns, Saurav Kadavath, Akul Arora, Steven Basart, Eric
  Tang, Dawn Song, and Jacob Steinhardt.
\newblock Measuring mathematical problem solving with the math dataset.
\newblock \emph{NeurIPS}, 2021.

\bibitem[Kojima et~al.(2023)Kojima, Gu, Reid, Matsuo, and
  Iwasawa]{lets_think_step_by_step_kojima2023largelanguagemodelszeroshot}
Takeshi Kojima, Shixiang~Shane Gu, Machel Reid, Yutaka Matsuo, and Yusuke
  Iwasawa.
\newblock Large language models are zero-shot reasoners, 2023.
\newblock URL \url{https://arxiv.org/abs/2205.11916}.

\bibitem[Leviathan et~al.(2022)Leviathan, Kalman, and
  Matias]{leviathan2022fastinferencetransformersspeculative}
Yaniv Leviathan, Matan Kalman, and Yossi Matias.
\newblock Fast inference from transformers via speculative decoding, 2022.
\newblock URL \url{https://arxiv.org/abs/2211.17192}.

\bibitem[Leviathan et~al.(2024)Leviathan, Kalman, and
  Matias]{leviathan2024selectiveattentionimprovestransformer}
Yaniv Leviathan, Matan Kalman, and Yossi Matias.
\newblock Selective attention improves transformer, 2024.
\newblock URL \url{https://arxiv.org/abs/2410.02703}.

\bibitem[McNemar(1947)]{mcnemar1947note}
Quinn McNemar.
\newblock Note on the sampling error of the difference between correlated
  proportions or percentages.
\newblock \emph{Psychometrika}, 12\penalty0 (2):\penalty0 153--157, 1947.

\bibitem[Mihaylov et~al.(2018)Mihaylov, Clark, Khot, and
  Sabharwal]{openbookqa_mihaylov2018suitarmorconductelectricity}
Todor Mihaylov, Peter Clark, Tushar Khot, and Ashish Sabharwal.
\newblock Can a suit of armor conduct electricity? a new dataset for open book
  question answering, 2018.
\newblock URL \url{https://arxiv.org/abs/1809.02789}.

\bibitem[OpenAI(2024)]{openai2024gpt4ocard}
OpenAI.
\newblock Gpt-4o system card, 2024.
\newblock URL \url{https://arxiv.org/abs/2410.21276}.

\bibitem[Raffel et~al.(2023)Raffel, Shazeer, Roberts, Lee, Narang, Matena,
  Zhou, Li, and Liu]{raffel2023exploringlimitstransferlearning}
Colin Raffel, Noam Shazeer, Adam Roberts, Katherine Lee, Sharan Narang, Michael
  Matena, Yanqi Zhou, Wei Li, and Peter~J. Liu.
\newblock Exploring the limits of transfer learning with a unified text-to-text
  transformer, 2023.
\newblock URL \url{https://arxiv.org/abs/1910.10683}.

\bibitem[Shaier(2024)]{shaier2024askingagainexploringllm}
Sagi Shaier.
\newblock Asking again and again: Exploring llm robustness to repeated
  questions, 2024.
\newblock URL \url{https://arxiv.org/abs/2412.07923}.

\bibitem[Springer et~al.(2024)Springer, Kotha, Fried, Neubig, and
  Raghunathan]{springer2024repetitionimproveslanguagemodel}
Jacob~Mitchell Springer, Suhas Kotha, Daniel Fried, Graham Neubig, and Aditi
  Raghunathan.
\newblock Repetition improves language model embeddings, 2024.
\newblock URL \url{https://arxiv.org/abs/2402.15449}.

\bibitem[Wang et~al.(2024)Wang, Ma, Zhang, Ni, Chandra, Guo, Ren, Arulraj, He,
  Jiang, Li, Ku, Wang, Zhuang, Fan, Yue, and
  Chen]{wang2024mmluprorobustchallengingmultitask}
Yubo Wang, Xueguang Ma, Ge~Zhang, Yuansheng Ni, Abhranil Chandra, Shiguang Guo,
  Weiming Ren, Aaran Arulraj, Xuan He, Ziyan Jiang, Tianle Li, Max Ku, Kai
  Wang, Alex Zhuang, Rongqi Fan, Xiang Yue, and Wenhu Chen.
\newblock Mmlu-pro: A more robust and challenging multi-task language
  understanding benchmark, 2024.
\newblock URL \url{https://arxiv.org/abs/2406.01574}.

\bibitem[Wei et~al.(2023)Wei, Wang, Schuurmans, Bosma, Ichter, Xia, Chi, Le,
  and Zhou]{wei2023chainofthoughtpromptingelicitsreasoning}
Jason Wei, Xuezhi Wang, Dale Schuurmans, Maarten Bosma, Brian Ichter, Fei Xia,
  Ed~Chi, Quoc Le, and Denny Zhou.
\newblock Chain-of-thought prompting elicits reasoning in large language
  models, 2023.
\newblock URL \url{https://arxiv.org/abs/2201.11903}.

\bibitem[Xu et~al.(2024)Xu, Tao, Shen, Xu, Xu, Long, guang Lou, and
  Ma]{xu2024rereadingimprovesreasoninglarge}
Xiaohan Xu, Chongyang Tao, Tao Shen, Can Xu, Hongbo Xu, Guodong Long, Jian
  guang Lou, and Shuai Ma.
\newblock Re-reading improves reasoning in large language models, 2024.
\newblock URL \url{https://arxiv.org/abs/2309.06275}.

\end{thebibliography}

\appendix

\newpage

\section{Appendix}

\subsection{Ablations and Variations}
\label{appendix:ablations and variations}

\begin{figure}[h!]
  \centering
  \includegraphics[width=1\textwidth]{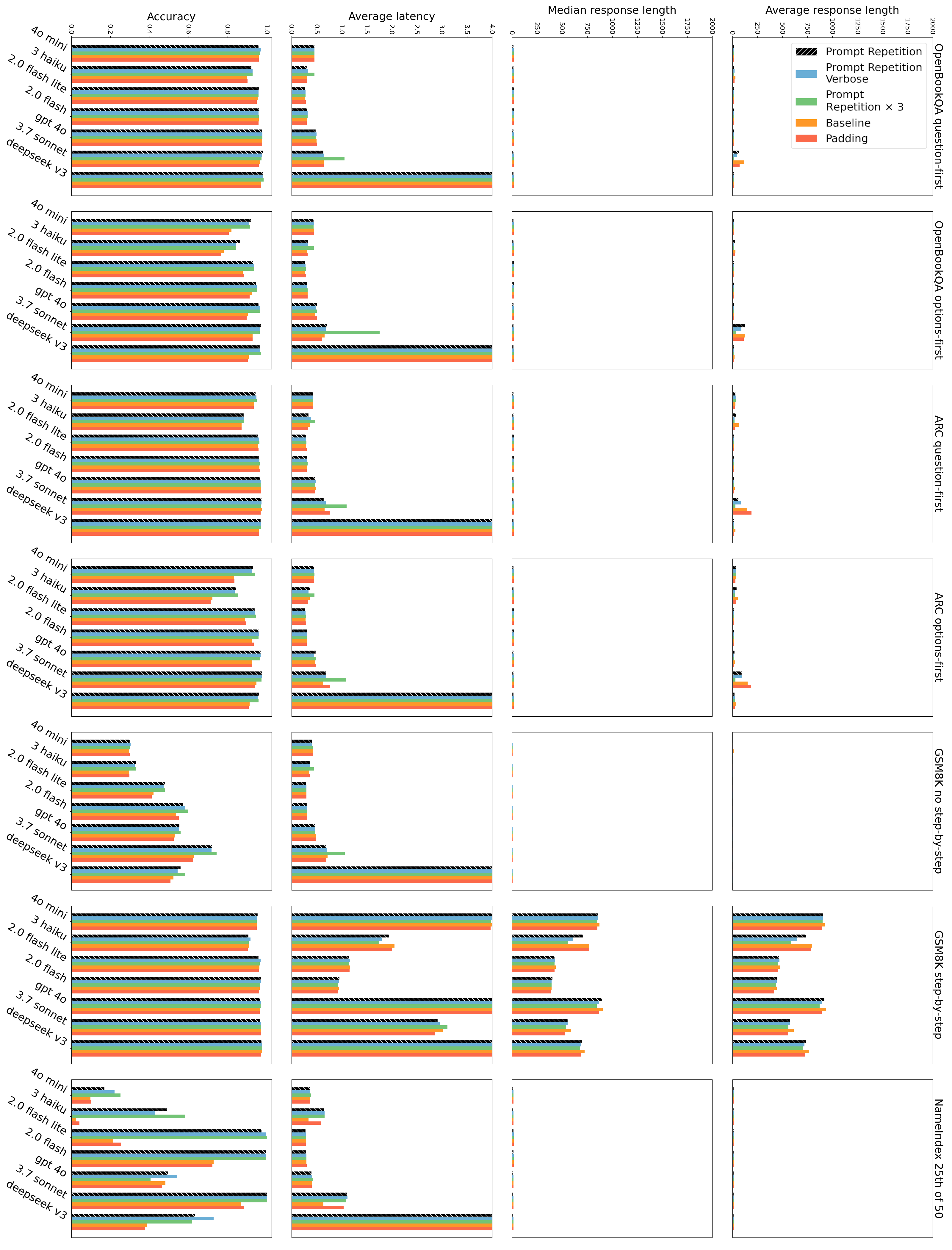}
  \caption{Comparison of accuracy, average, and median response length, as well as average latency, across methods and benchmarks (1).}
  \label{fig:ablations}
\end{figure}

\newpage

\begin{figure}[h!]
  \centering
  \includegraphics[width=1\textwidth]{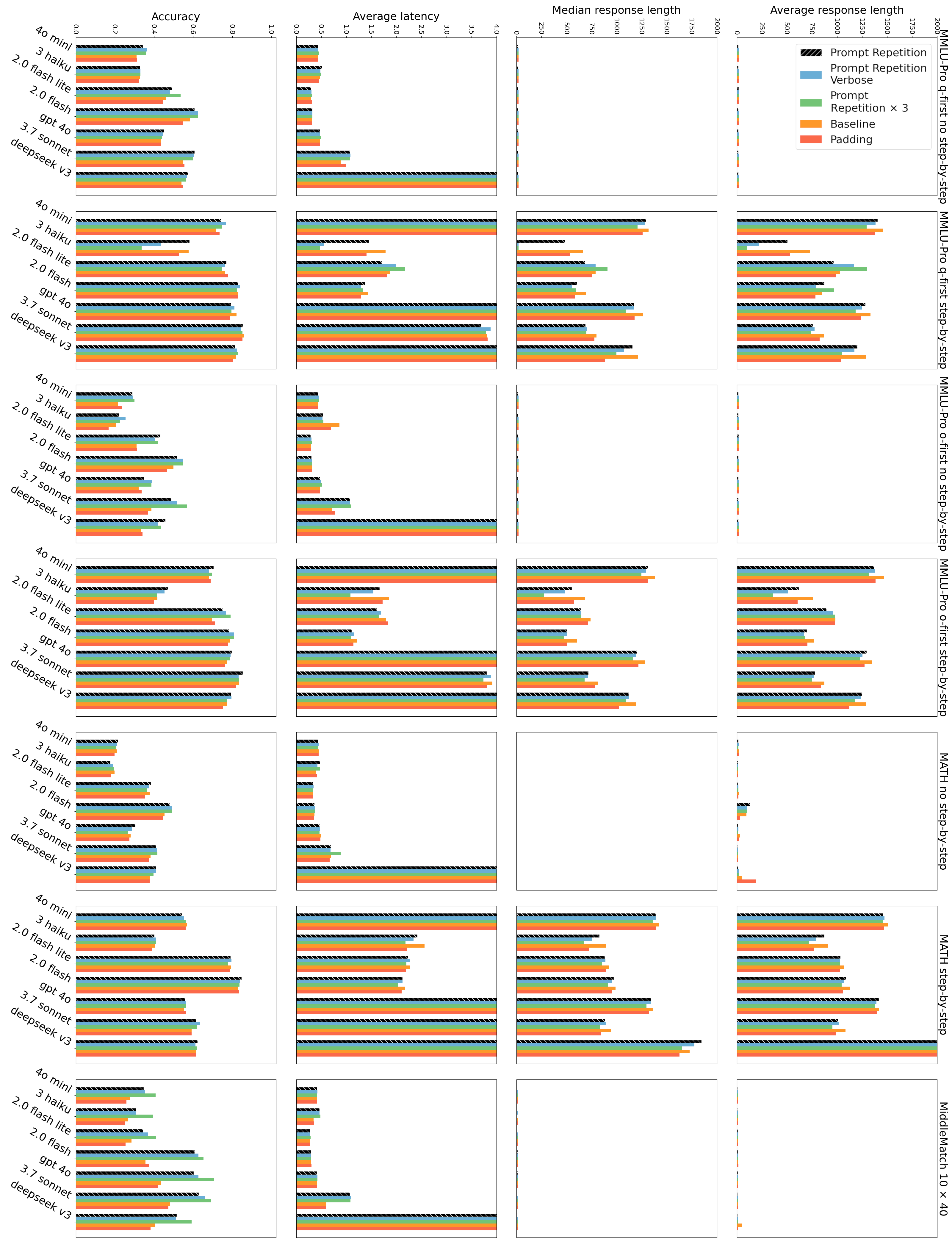}
  \caption{Comparison of accuracy, average, and median response length, as well as average latency, across methods and benchmarks (2).}
  \label{fig:ablations2}
\end{figure}

\newpage

\subsection{Prompt Repetition with Reasoning}
\label{appendix:with reasoning}

When asking the models to think step by step, we observe that prompt repetition is neutral to slightly positive (5 wins, 1 loss, 22 ties), which is expected given that the reasoning often starts with repeating (parts-of) the prompt anyway.

\begin{figure}[h]
  \centering
  \begin{adjustbox}{center}
    \includegraphics[width=\textwidth]{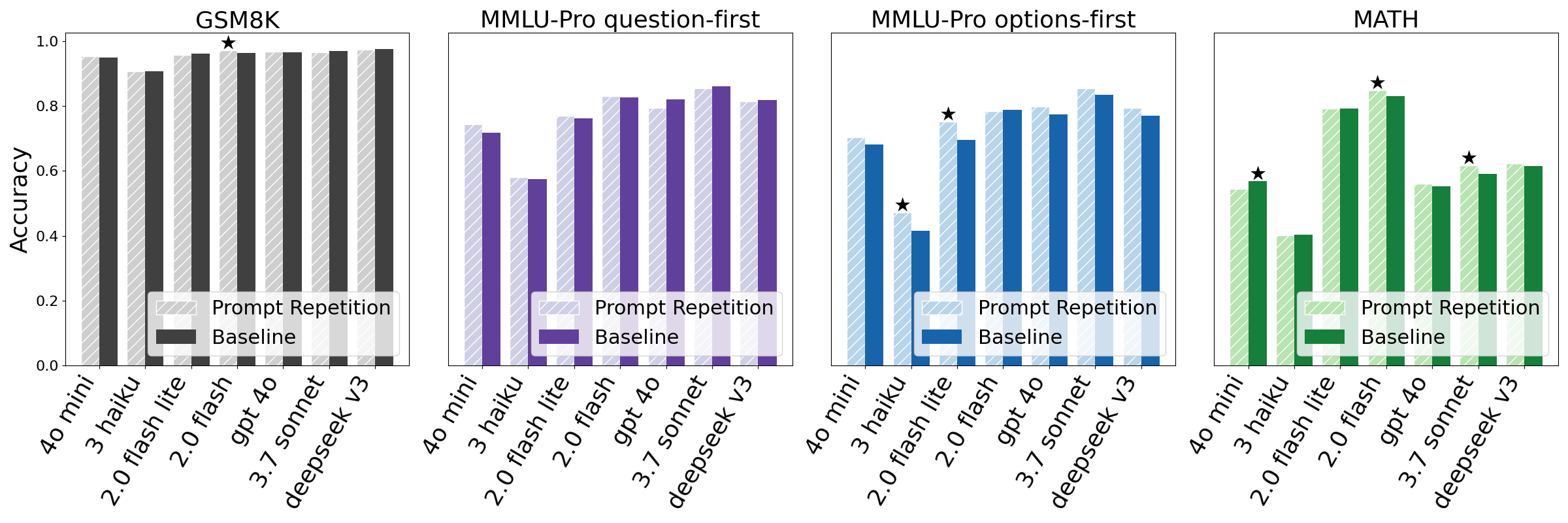}
  \end{adjustbox}
  \caption{Prompt repetition vs. baseline accuracy for popular LLMs and various benchmarks when asking the model to think step by step. A star indicates a statistically significant win ($p_{value} < 0.1$ according to the McNemar test \cite{mcnemar1947note}). Prompt repetition wins 5 out of 28 tests, with 1 loss.}
  \label{fig:sbs}
\end{figure}

\newpage

\subsection{Custom Tasks}
\label{appendix-custom-tasks}

In addition to the standard benchmarks, we also evaluate prompt repetition on two custom tasks specifically designed to demonstrate its usefulness.

\paragraph{NameIndex} Here the model gets a list of $N$ names and is asked to output the $i$th name on the list. We use $N=50, i=25$. Example:

\textcolor{gray}{
\texttt{Here's a list of names:\\
\\
Dale Lopez, Peter Sanchez, Allen Harris, Scott Davis, Hudson Leviathan, Daphne Kalman, Dennis Davis, Henry King, Alfred Cooper, Bruce Usher, Travis Ramirez, Rafael Jennings, Richard Rogers, Walter Young, Caleb Harris, Ben Kalman, Donald Carter, Richard Sterling, Mark Nightingale, Steven Carter, Talia Kalman, Dennis Hanson, James Harris, Craig Chavez, Paul Sanchez, Samuel Curtis, Jacob James, Allen Thomas, Dale Evans, James Fox, Douglas Allen, Orion Johnson, Alexander Wright, Eugene Morrison, Nelson Lee, Alan Young, Caleb Ward, Alberto Robinson, Robert McCarthy, Mark Price, Kenneth Ramirez, Jeffrey White, Chad Cooper, Arthur Waters, Bruce Callahan, Liam Leviathan, Steven Robinson, Alberto Murphy, Leonard Johnson, Robert Murphy\\
\\
What's the 25th name?}}

\paragraph{MiddleMatch} Here the model gets a list of $N$ names or numbers (out of a total possible set of $K$, i.e., $K < N$ means there will be repeating elements), and is asked to output the name/number located directly between two given ones. We use $N=40, K=10$. Example:

\textcolor{gray}{
\texttt{Here's a list (potentially with repetitions) of names:\\
\\
Carlos Davis, Dale Sims, Carlos Davis, Dale Sims, Stephen Cruz, Dale Sims, Finnian Ross, Stephen Cruz, Stephen Cruz, Gregory Collins, Dale Sims, Stephen Cruz, Carlos Davis, Stephen Cruz, Dale Sims, Dale Sims, Stephen Cruz, Stephen Cruz, Leonard Kalman, Bruce Phillips, Raymond Roberts, Dale White, Leonard Kalman, Finnian Ross, James Wright, Finnian Ross, Raymond Roberts, Dale Sims, Dale Sims, Leonard Kalman, Dale Sims, Carlos Davis, Leonard Kalman, Bruce Phillips, Dale Sims, Raymond Roberts, Gregory Collins, Gregory Collins, Dale Sims, Finnian Ross\\
\\
What is the single name that appears right between Carlos Davis and Bruce Phillips?}}

\newpage

\subsection{Query Examples}
\label{appendix-query-examples}

\begin{longtable}{ p{.1\textwidth} | p{.3\textwidth} | p{.55\textwidth} }

Method & Template & Example Query \\
\hline
\endfirsthead

\multicolumn{3}{c}{\textit{(Continued from previous page)}} \\
\noalign{\vspace{2ex}}
Method & Template & Example Query \\
\hline
\endhead

\noalign{\vspace{2ex}}
\multicolumn{3}{c}{\textit{(Continued on next page)}} \\
\endfoot

\hline
\endlastfoot

Baseline & \texttt{\makecell[l]{<QUERY>}} & \makecell[l]{ \\ Which of the following combinations is a mixture rather \\ than a compound? \\  \\ A. oxygen and nitrogen in air \\ B. sodium and chlorine in salt \\ C. hydrogen and oxygen in water \\ D. nitrogen and hydrogen in ammonia \\  \\ Reply with one letter ('A', 'B', 'C', 'D') in the format: \\ The answer is <ANSWER>. \\ \\ } \\
\hline
\makecell[l]{Prompt \\ Repetition} & \texttt{\makecell[l]{<QUERY> \\  \\ <QUERY>}} & \makecell[l]{ \\ Which of the following combinations is a mixture rather \\ than a compound? \\  \\ A. oxygen and nitrogen in air \\ B. sodium and chlorine in salt \\ C. hydrogen and oxygen in water \\ D. nitrogen and hydrogen in ammonia \\  \\ Reply with one letter ('A', 'B', 'C', 'D') in the format: \\ The answer is <ANSWER>. \\  \\ Which of the following combinations is a mixture rather \\ than a compound? \\  \\ A. oxygen and nitrogen in air \\ B. sodium and chlorine in salt \\ C. hydrogen and oxygen in water \\ D. nitrogen and hydrogen in ammonia \\  \\ Reply with one letter ('A', 'B', 'C', 'D') in the format: \\ The answer is <ANSWER>. \\ \\ } \\
\hline
\makecell[l]{Prompt \\ Repetition \\ (Verbose)} & \texttt{\makecell[l]{<QUERY> \\  \\ Let me repeat that: \\  \\ <QUERY>}} & \makecell[l]{ \\ Which of the following combinations is a mixture rather \\ than a compound? \\  \\ A. oxygen and nitrogen in air \\ B. sodium and chlorine in salt \\ C. hydrogen and oxygen in water \\ D. nitrogen and hydrogen in ammonia \\  \\ Reply with one letter ('A', 'B', 'C', 'D') in the format: \\ The answer is <ANSWER>. \\  \\ Let me repeat that: \\  \\ Which of the following combinations is a mixture rather \\ than a compound? \\  \\ A. oxygen and nitrogen in air \\ B. sodium and chlorine in salt \\ C. hydrogen and oxygen in water \\ D. nitrogen and hydrogen in ammonia \\  \\ Reply with one letter ('A', 'B', 'C', 'D') in the format: \\ The answer is <ANSWER>. \\ \\ } \\
\hline
\makecell[l]{Prompt \\ Repetition \\ $\times 3$ } & \texttt{\makecell[l]{<QUERY> \\  \\ Let me repeat that: \\  \\ <QUERY> \\  \\ Let me repeat that one \\ more time: \\  \\ <QUERY>}} & \makecell[l]{ \\ Which of the following combinations is a mixture rather \\ than a compound? \\  \\ A. oxygen and nitrogen in air \\ B. sodium and chlorine in salt \\ C. hydrogen and oxygen in water \\ D. nitrogen and hydrogen in ammonia \\  \\ Reply with one letter ('A', 'B', 'C', 'D') in the format: \\ The answer is <ANSWER>. \\  \\ Let me repeat that: \\  \\ Which of the following combinations is a mixture rather \\ than a compound? \\  \\ A. oxygen and nitrogen in air \\ B. sodium and chlorine in salt \\ C. hydrogen and oxygen in water \\ D. nitrogen and hydrogen in ammonia \\  \\ Reply with one letter ('A', 'B', 'C', 'D') in the format: \\ The answer is <ANSWER>. \\  \\ Let me repeat that one more time: \\  \\ Which of the following combinations is a mixture rather \\ than a compound? \\  \\ A. oxygen and nitrogen in air \\ B. sodium and chlorine in salt \\ C. hydrogen and oxygen in water \\ D. nitrogen and hydrogen in ammonia \\  \\ Reply with one letter ('A', 'B', 'C', 'D') in the format: \\ The answer is <ANSWER>. \\ \\ } \\
\hline
Padding & \texttt{\makecell[l]{<QUERY> \\  \\ Ignore these periods \\ (they are irrelevant) \\ and answer the above \\ question: ....... \\
...<LEN(QUERY)>...}} & \makecell[l]{ \\ Which of the following combinations is a mixture rather \\ than a compound? \\  \\ A. oxygen and nitrogen in air \\ B. sodium and chlorine in salt \\ C. hydrogen and oxygen in water \\ D. nitrogen and hydrogen in ammonia \\  \\ Reply with one letter ('A', 'B', 'C', 'D') in the format: \\ The answer is <ANSWER>.. \\  \\   Ignore these periods (they are irrelevant) and answer the \\ above question: ......................................................................\\................................................................................................\\................................................................................................\\....................... \\ \\ } \\

\end{longtable}

\end{document}